\documentclass[10pt,twocolumn,letterpaper]{article}

\usepackage{iccv}
\usepackage{times}
\usepackage{epsfig}
\usepackage{graphicx}
\usepackage{amsmath}
\usepackage{amssymb}
\usepackage{multirow}

\usepackage{enumitem}

\usepackage[pagebackref=true,breaklinks=true,letterpaper=true,colorlinks,bookmarks=false]{hyperref}

\iccvfinalcopy 

\def\iccvPaperID{178} 

\ificcvfinal\fi
\begin{document}

\title{Image Synthesis From Reconfigurable Layout and Style}
 
\author{Wei Sun and Tianfu Wu\\
Department of ECE and the Visual Narrative Initiative, North Carolina State University\\
{\tt\small\{wsun12, tianfu\_wu\} @ncsu.edu}
}

\maketitle

\begin{abstract}
   Despite remarkable recent progress on both unconditional and conditional image synthesis, it remains a long-standing problem to learn generative models that are capable of synthesizing realistic and sharp images from reconfigurable spatial layout (i.e., bounding boxes + class labels in an image lattice) and style (i.e., structural and appearance variations encoded by latent vectors), especially at high resolution. By reconfigurable, it means that a model can preserve the intrinsic one-to-many mapping from a given  layout to multiple plausible images with different styles, and is adaptive with respect to perturbations of a layout and style latent code. In this paper, we present a layout- and style-based architecture for generative adversarial networks (termed LostGANs) that can be trained end-to-end to generate images from reconfigurable layout and style. Inspired by the vanilla StyleGAN, the proposed LostGAN consists of two new components:  (i) learning fine-grained mask maps in a weakly-supervised manner to bridge the gap between layouts and images, and (ii) learning object instance-specific layout-aware feature normalization (ISLA-Norm) in the generator to realize multi-object style generation. In experiments, the proposed method is tested on the COCO-Stuff dataset and the Visual Genome dataset with state-of-the-art performance obtained. The code and pretrained models are available at \url{https://github.com/iVMCL/LostGANs}.       
\end{abstract}

\section{Introduction}
\vspace{-2mm}
\subsection{Motivation and Objective}
\vspace{-2mm}
Remarkable recent progress has been made on both unconditional and conditional image synthesis~\cite{goodfellow2014generative,radford2015unsupervised,zhang2018self,miyato2018spectral,brock2018large,miyato2018cgans,karras2017progressive,karras2018style}. The former aims to generate high-fidelity images from some random latent codes. The latter needs to do so with given conditions satisfied in terms of some consistency metrics. The conditions may take many forms such as categorical labels, desired attributes, descriptive sentences, scene graphs, and paired or unpaired images/semantic maps. From the perspective of generative learning, the solution space of the latter is much difficult to capture than that of the former. Conditional image synthesis, especially with coarse yet complicated and reconfigurable conditions, remains a long-standing problem. Once powerful systems are developed, they can facilitate to pave a way for computers to truly understand visual patterns via analysis-by-synthesis. They will also enable a wide range of practical applications, e.g., generating high-fidelity data for long-tail scenarios in different vision tasks such as autonomous driving. 

\begin{figure}
    \centering
    \includegraphics[width=1.0\linewidth]{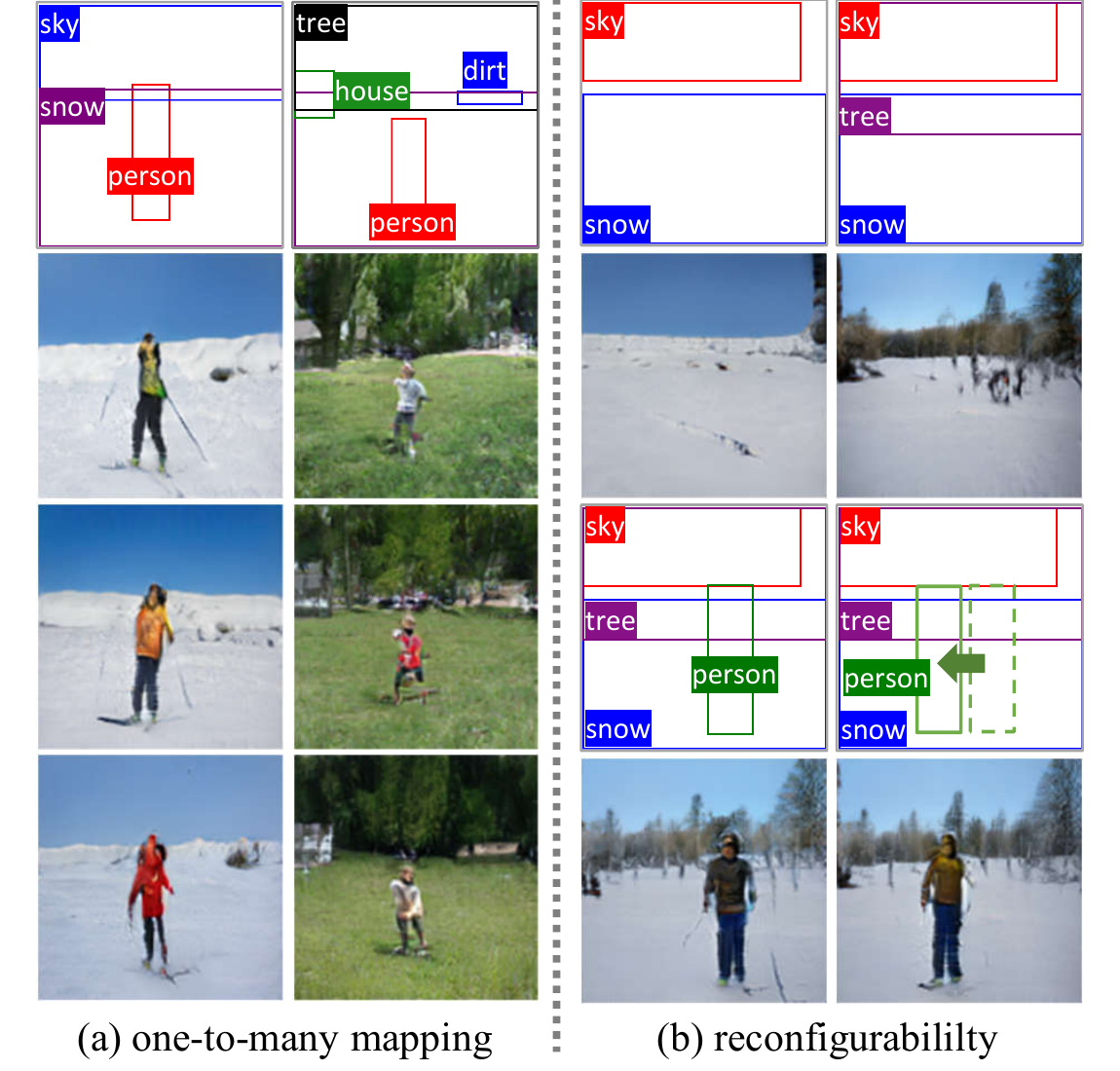}
    \caption{Illustration of the proposed method. \textit{Left:} Our model preserves one-to-many mapping for image synthesis from layout and style. Three samples are generated for each input layout by sampling the style latent codes. \textit{Right:} Our model is also adaptive w.r.t. reconfigurations of layouts (by adding new object bounding boxes or changing the location of a bounding box). The results are generated at resolution $128\times 128$. See text for details.}
    \label{fig:teaser}
    \vspace{-4mm}
\end{figure}

\begin{figure*} [ht]
    \centering
    \includegraphics[width=0.95\linewidth]{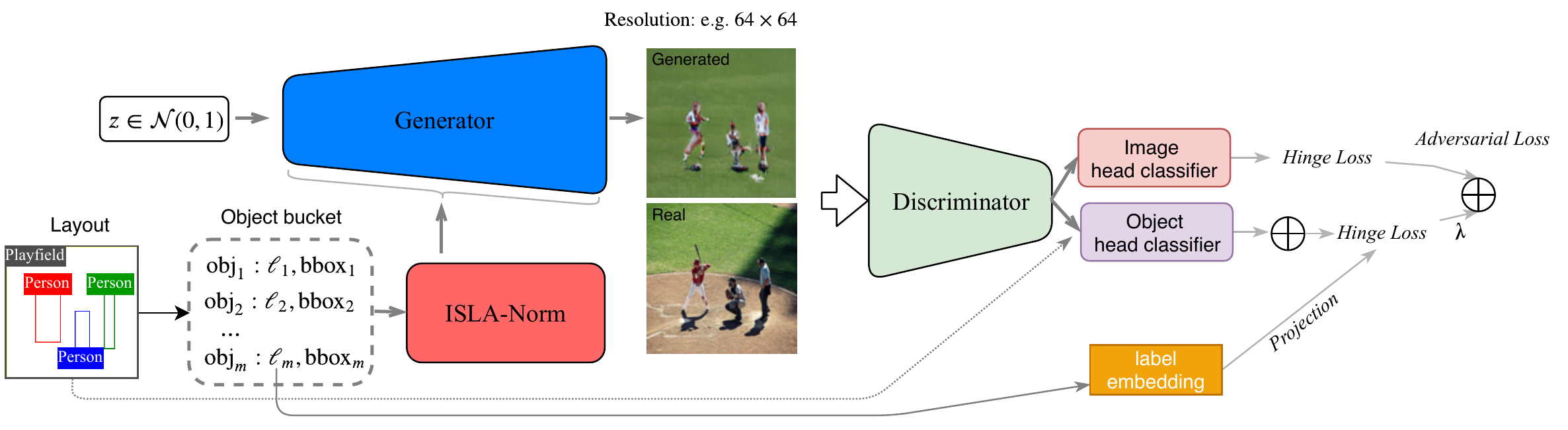}
    \caption{Illustration of the proposed layout- and style-based GANs (LostGANs) for image synthesis from reconfigurable layout and style. Both the generator and discriminator use ResNets as backbones. See text for details. }
    \label{fig:workflow}
    \vspace{-4mm}
\end{figure*}

\begin{figure} [!ht]
    \centering
    \includegraphics[width=1.0\linewidth]{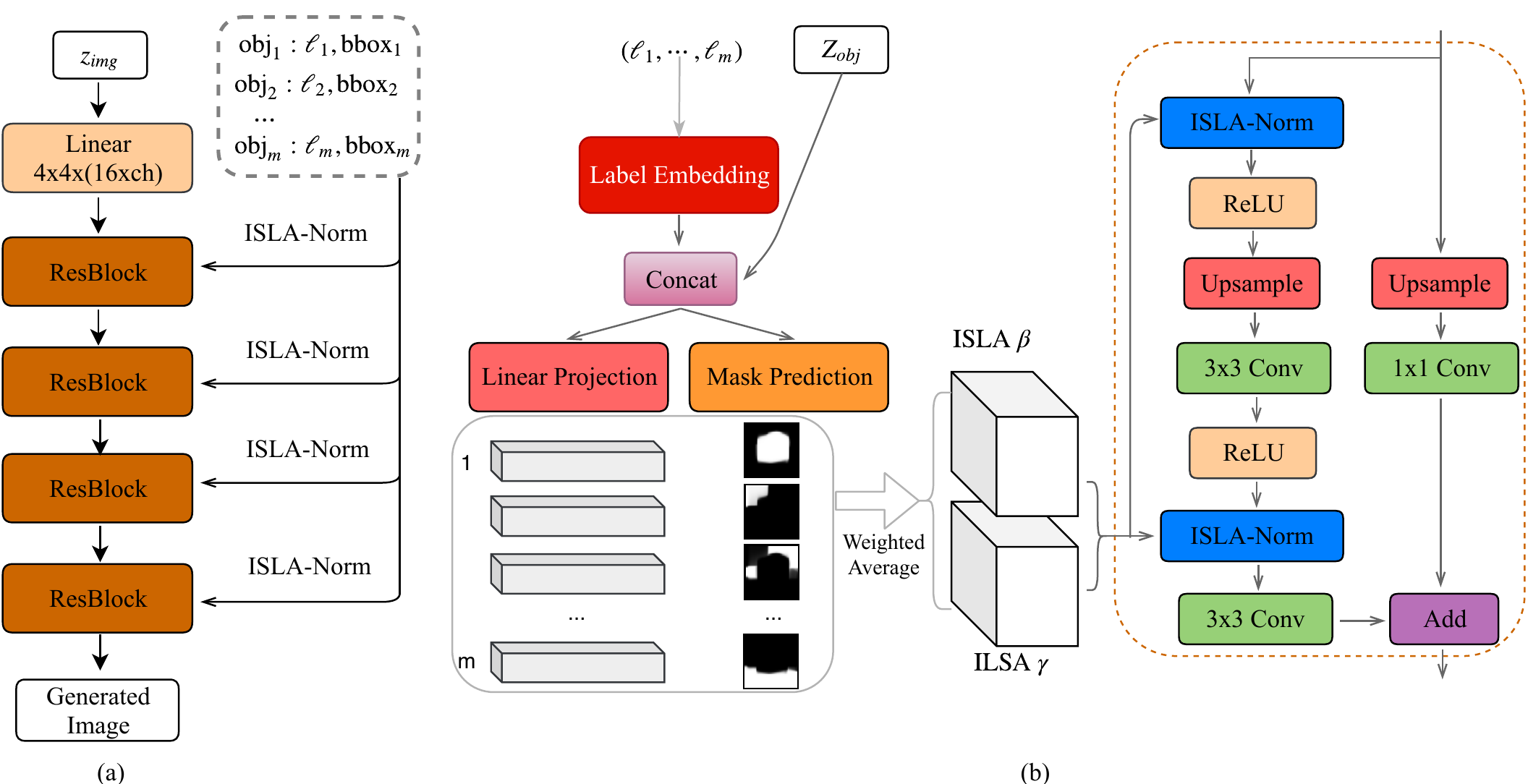}
    \caption{Illustration of the generator (a) and the ISLA-Norm (b) in our LostGAN. See text for details. Best viewed in magnification.}
    \label{fig:LostGAN}\vspace{-4mm}
\end{figure}

In this paper, we are interested in conditional image synthesis from layout and style. The layout consists of labeled bounding boxes configured in an image lattice (e.g., $64\times 64$ or $128\times 128$). The style is represented by some latent code. Layout represents a sweet yet challenging spot for conditional image synthesis: First, layout is usually used as the intermediate representation for other conditional image synthesis such as text-to-image~\cite{zhang2017stackgan,xu2018attngan} and scene-graph-to-image~\cite{johnson2018image}. Second, layout is more flexible, less constrained and easier to collect than semantic segmentation maps~\cite{isola2017image,wang2018high}. Third, layout-to-image requires addressing challenging one-to-many mapping and consistent multi-object generation (e.g., occlusion handling for overlapped bounding boxes and uneven, especially long-tail distributions of objects).  

Layout-to-image is a relatively new task with many new technical challenges for state-of-the-art image synthesis frameworks and only a few work have been proposed in the very recent literature~\cite{johnson2018image,hong2018inferring,zhao2018image}. 
Recently, we have seen remarkable progress on the high-fidelity class-conditional image synthesis in ImageNet by the BigGAN~\cite{brock2018large}, and on the amazing style control for specific objects (e.g., faces and cars) by the unconditional StyleGAN \cite{karras2018style} (which may be considered as implicitly conditional image synthesis since only one category is usually handled in training). Despite the big successes in generative learning, the problem considered in this paper is still more challenging since the solution space is much more difficult to capture and has much more complicated distributions. For example, we can use the BigGAN to generate a cat image, and as long as the generated image looks realistic and sharp, we think it does a great job. Similarly, we can use the StyleGAN to generate a face image, and we are happy (even shocked sometimes) if a realistic and sharp face image is generated with a natural style (e.g., smile or sad). Layout-to-image needs to tackle many spatial and semantic (combinatorial) relationships among multiple objects besides the naturalness. 

In this paper, we further focus on image synthesis from \textit{reconfigurable} layout and style. By reconfigurable, it means that a model can preserve the intrinsic one-to-many mapping from a given layout to multiple plausible images with different styles, and is adaptive with respect to perturbations of layout and style latent code (Figure~\ref{fig:teaser}). State-of-the-art methods on reconfigurable layout-to-image still mainly focus on low resolution ($64\times 64$)~\cite{johnson2018image,zhao2018image} (which are, in part, due to  computationally expensive designs in the pipelines such as convolutional LSTM used in~\cite{zhao2018image}). Beside the resolution issue, another drawback of existing methods is that the diversity of generated images (i.e., style control) is not sufficiently high to preserve the intrinsic one-to-many mapping. We aim to improve both the resolution and the style diversity in reconfigurable layout-to-image.

\subsection{Method Overview}\vspace{-2mm}
To address the challenges in layout-to-image and inspired by the recent StyleGANs~\cite{karras2018style}, we present a \textbf{L}ay\textbf{O}ut- and \textbf{ST}yle-based architecture for GANs (termed \textit{LostGANs}) in the paper (Figure~\ref{fig:workflow}).

First, since layout-to-image entails highly expressive neural architectures handling multi-object generation and their diverse occurrence and configurations in layouts. We utilize ResNet~\cite{he2016deep} for both the generator and discriminator in the proposed LostGAN, as done in the projection-based cGAN~\cite{miyato2018cgans} and BigGAN~\cite{brock2018large}.

Second, to account for the gap between bounding boxes in a layout and underlying object shapes, we introduce an encoder for layout to predict masks for each bounding box. As we will show in experiments, our LostGAN can predict reasonably good masks in a weakly-supervised manner. The masks help place objects in the generated images with fine-grained geometric properties. So, \textbf{we address layout-to-image by computing layout-to-mask-to-image} (Figure~\ref{fig:LostGAN}), which is motivated by impressive recent progress on conditional image synthesis from semantic label maps~\cite{isola2017image,wang2018high}. 

Third, to achieve instance-sensitive and layout-aware style control, we extend the Adaptive Instance Normalization (AdaIN) used in the StyleGAN~\cite{karras2018style} to \textbf{object instance-specific and layout-aware feature normalization (ISLA-Norm)} for the generator for fine-grained spatially distributed multi-object style control. ISLA-Norm computes the mean and variance as done in BatchNorm~\cite{ioffe2015batch}, but computes object instance-specific and layout-aware affine transformations (i.e., gamma and beta parameters) separately for each sample in a min-batch as done in AdaIN (Figure~\ref{fig:LostGAN}). We utilize the projection-based approach proposed in~\cite{brock2018large}. From the layout encoder, we compute object instance-specific style latent codes (gamma and beta parameters) via simple linear projection. Then, we place the projection-based latent codes in the corresponding predicted masks, and thus induce layout-aware affine transformations for recalibrating normalized feature responses. 

Lastly, we utilize both image and object adversarial hinge losses~\cite{HingeLoss1,HingeLoss2} as adopted in~\cite{miyato2018spectral,miyato2018cgans} in the end-to-end training. Object adversarial loss follows the projection based method in~\cite{miyato2018cgans} which is the state-of-the-art approach for embedding labels. 

We deliberately try to keep our LostGAN as simple as possible by exploiting the best practices in the literature of conditional image synthesis. We hope it can stimulate more exploration on this relatively new task, image synthesis from reconfigurable layout and style. 

In experiments, our LostGAN is tested in the COCO-Stuff dataset~\cite{caesar2018coco} and the Visual Genome (VG) dataset~\cite{krishna2017visual}. It obtains state-of-the-art performance on both datasets in terms of the inception score~\cite{salimans2016improved},  Fr\`echet Inception Distance~\cite{FID}, diversity score~\cite{zhang2018unreasonable}, and classification accuracy~\cite{ravuri2019classification}, which supports the effectiveness of our ILSA-Norm and LostGAN.

\section{Related Works}\vspace{-2mm}
\textbf{Conditional Image Synthesis.}
Generative Adversarial Networks (GANs) \cite{goodfellow2014generative} have achieved great success in image synthesis conditioned on additional input information (i.e. class information \cite{odena2017conditional,miyato2018cgans,zhang2018self}, source image \cite{kim2017learning,zhu2017unpaired,huang2018multimodal}, text description \cite{reed2016generative,zhang2017stackgan}, etc). How to feed conditional information to model has been studied in various ways. In \cite{odena2017conditional,reed2016generative} vector encoded from conditional information concatenated with noise vector is passed as input to generator. In \cite{de2017modulating,dumoulin2017learned,brock2018large,park2019semantic}, conditional information is provided to generator by conditional gains and bias in BatchNorm \cite{ioffe2015batch} layers. Concurrent work \cite{park2019semantic} learns spatially adaptive normalization from well annotated semantic masks, while our proposed ISLA-Norm learns from coarse layout information. \cite{reed2016generative,dumoulin2016adversarially,zhang2017stackgan} feed the conditional information into discriminator by naively concatenation with the input or intermediate feature vector. In \cite{miyato2018cgans}, projection based way to incorporate conditional information to discriminator effectively improve the quality of class conditional image generation. In our proposed method, layout condition is adopted to generator with ISLA-Norm, and objects information is utilized in projection based discriminator as \cite{miyato2018cgans}.

\textbf{Image Synthesis from Layout.}
Spatial layout conditioned image generation has been studied in recent literature. In \cite{johnson2018image,hong2018inferring,hinz2019generating,li2019object}, layout and object information is utilized in text-to-image generation. \cite{hinz2019generating} controls location of multiple objects in text-to-image generation by adding an object pathway to both the generator and discriminator. \cite{johnson2018image,hong2018inferring,li2019object} performs text-to-image synthesis in two steps: semantic layout (class label and bounding boxes) generation from text first, and image synthesis conditioned on predicted semantic layout and text description. However, \cite{hong2018inferring,li2019object} requires pixel-level instance segmentation annotation, which is labor intensive to collect, for training of shape generator, while our method does not require pixel-level annotation and can learn segmentation mask in a weakly-supervised manner. \cite{zhao2018image} studied similar task with us, where variational autoencoders based network is adopted for scene image generation from layout.

\noindent\textbf{Our Contributions.} This paper makes the following main contributions to the field of conditional image synthesis. 
\begin{itemize} [leftmargin=*]
\itemsep0em
    \item It presents a layout- and style-based architecture for GANs (termed LostGANs) which integrates the best practices in conditional and unconditional GANs for a relatively new task, image synthesis from reconfigurable layout and style. 
    \item It presents an object instance-specific and layout-aware feature normalization scheme (termed ISLA-Norm) which is inspired by the projection-based conditional BatchNorm used in cGANs~\cite{brock2018large} and the Adaptive Instance Normalization (AdaIN) used in StyleGAN~\cite{karras2018style}. It explicitly accounts for the layout information in the affine transformations.  
    \item It shows state-of-the-art performance in terms of the inception score~\cite{salimans2016improved}, Fr\`echet Inception Distance~\cite{FID}, diversity score~\cite{zhang2018unreasonable} and classification accuracy~\cite{ravuri2019classification} on two widely used datasets, the COCO-Stuff~\cite{caesar2018coco} and the Visual Genome~\cite{krishna2017visual}. 
\end{itemize}

\section{The Proposed Method}\vspace{-2mm}
In this section, we first define the problem and then present details of our LostGAN and ISLA-Norm. 

\subsection{Problem Formulation}\vspace{-2mm}
Denote by $\Lambda$ an image lattice (e.g., $64\times 64$). Let $L=\{(\ell_i, bbox_i)_{i=1}^m\}$ be a layout consisting of $n$ labeled bounding boxes, where label $\ell_i\in \mathcal{C}$ (e.g., $|\mathcal{C}|=171$ in the COCO-Stuff dataset), and bounding box $bbox_i\subseteq \Lambda$. Different bounding boxes may have occlusions. Let $z_{img}$ be the latent code controlling image style and $z_{obj_i}$ the latent code controlling object instance style for $(\ell_i, bbox_i)$ (e.g., the latent codes are sampled from the standard normal distribution, $\mathcal{N}(0,1)$ under i.i.d. setting). Denote by $Z_{obj}=\{z_{obj_i}\}_{i=1}^m$ the set of object instance style latent codes.

Image synthesis from layout and style is the problem of learning a generation function which is capable of synthesizing an image defined on $\lambda$ for a given input $(L, z_{img}, Z_{obj})$,
\begin{equation}
    I = G(L, z_{img}, Z_{obj}; \Theta_G) \label{eq:generator}
\end{equation}
where $\Theta_G$ represents the parameters of the generation function. Ideally, $G(\cdot)$ is expected to capture the underlying conditional data distribution $p(I|L, z_{img}, Z_{obj})$ in the high-dimensional space.

\textbf{Reconfigurability of $G(\cdot)$.} We are interested in three aspects in this paper:
\begin{itemize}[leftmargin=*]
\itemsep0em
    \item \textit{Image style reconfiguration:} If we fix the layout $L$, is $G(\cdot)$ capable of generating images with different styles for different $(z_{img}, Z_{obj})$?
    \item \textit{Object style reconfiguration:} If we fix the layout $L$, the image style $z_{img}$ and object styles $Z_{obj}\setminus z_{obj_i}$,  is $G(\cdot)$ capable of generating consistent images with different styles for the object  $(\ell_i, bbox_i)$ using different $z_{obj_i}$?
    \item \textit{Layout reconfiguration:} Given a $(L, z_{img}, Z_{obj})$, is $G(\cdot)$ capable of generating consistent images for different $(L^+, z_{img}, Z_{obj}^+)$ where we can add a new object to $L^+$ or just change the bounding box location of an existing object? When a new object is added, we also sample a new $z_{obj}$ to add in $Z_{obj}^+$.  
\end{itemize}
It is a big challenge to address the three aspects by learning a single generation function. It may be even difficult for well-trained artistic people to do so at scale (e.g., handling the 171 categories in the COCO-Stuff dataset). Due to the complexity that the generation function (Eqn.~\ref{eq:generator}) needs to handle, it is parameterized (often over-parameterized) by powerful deep neural networks (DNNs). It is also well-known that training the DNN-based generation function individually is a extremely difficult task. Generative adversarial networks (GANs)~\cite{goodfellow2014generative} are entailed which are formulated under two-player minmax game settings. 

\subsection{The LostGAN}\vspace{-2mm}
As Figure~\ref{fig:workflow} shows, our LostGAN follows the traditional GAN pipeline with the following modifications. 

\vspace{-2mm}
\subsubsection{The Generator}\vspace{-2mm}
Figure~\ref{fig:LostGAN} (a) shows the generator which utilizes the ResNet~\cite{he2016deep} architecture as backbone. Consider generating $64\times 64$ images, the generator consists of 4 residual building blocks (ResBlocks). The image style latent code $z_{img}$ is a $d_{noise}$-dim vector ($d_{noise}=128$ in our experiments) whose elements are sampled from standard normal distribution under i.i.d. setting. Through a linear fully connected (FC) layer, $z_{img}$ is projected to a $4\times 4\times (16\times ch)$ dimensional vector which is then reshaped to $(4, 4, 16\times ch)$ (representing height, width and channels) where $ch$ is a hyperparameter to control model complexity (e.g., $ch=64$ for generating $64\times 64$ images). Then, each of the four ResBlocks upsamples its input with ratio $2$ and bilinear interpolation. In the meanwhile, the feature channel will be decreased by ratio $2$. For generating $128\times 128$ images, we use 5 ResBlocks with $ch=64$ and the same $d_{noise}=128$ for $z_{img}$. 
\vspace{-2mm}
\subsubsection{The ISLA-Norm}\vspace{-2mm}
Figure~\ref{fig:LostGAN} (b) shows the detail of ResBlock and the proposed ISLA-Norm. The ResBlock uses the basic block design as adopted in the projection-based cGAN~\cite{miyato2018cgans} and BigGAN~\cite{brock2018large}. Our ISLA-Norm first computes the mean and variance as done in BatchNorm~\cite{ioffe2015batch}, and then learns object instance-specific layout-aware affine transformation for each sample in a batch similar in spirit to the AdaIN used by the StyleGAN~\cite{karras2018style}. So, the feature normalization is computed in a batch manner, and the affine transformation is recalibrated in a sample-specific manner. 

Denote by $x$ the input 4D feature map of ISLA-Norm, and $x_{nhwc}$ the feature response at position $(n, h, w, c)$ (using the convention order of axes for batch, spatial height and width axis, and channel). We have $n\in [0, N-1], \, h\in [0, H-1],\, w\in [0, W-1]$ and $c\in [0, C-1]$ where $H, W, C$ depend on the stage of a ResBlock. 

In training, ISLA-Norm first normalizes $x_{nhwc}$ by,
\begin{equation}
    \hat{x}_{nhwc} = \frac{x_{nhwc} - \mu_c}{\sigma_c},
\end{equation}
where the channel-wise batch mean $\mu_c = \frac{1}{N\cdot H\cdot W} \sum_{n,h,w} x_{nhwc}$ and  standard deviation (std) $\sigma_c=\sqrt{\frac{1}{N\cdot H\cdot W} \sum_{n,h,w} (x_{nhwc}-\mu_c)^2 +\epsilon}$ ($\epsilon$ is a small positive constant for numeric stability). In standard BatchNorm~\cite{ioffe2015batch}, for the affine transformation, a channel-wise $\gamma_c$ and $\beta_c$ will be learned and shared with all spatial locations and all samples in a batch. our ISLA-Norm will learn object instance-specific and layout-aware affine transformation parameters, $\gamma_{nhwc}$ and $\beta_{nhwc}$, and then recalibrate the normalized feature responses by, 
\begin{equation}
    \Tilde{x}_{nhwc} = \gamma_{nhwc}\cdot \hat{x}_{nhwc} + \beta_{nhwc}. 
\end{equation}

\textbf{Computing $\gamma_{nhwc}$ and $\beta_{nhwc}$}. Without loss of generality, we show how to compute the gamma and beta parameters for one sample, i.e., $\gamma_{hwc}$ and $\beta_{hwc}$. As shown in Figure~\ref{fig:LostGAN} (b), we have the following four steps. 

\textit{i) Label Embedding.} We use one-hot label vector for the $m$ object instances and then we obtain the $m\times d_{\ell}$ one-hot label matrix (e.g., $d_{\ell}=171$ in COCO-Stuff). For label embedding, we use a learnable $d_{\ell}\times d_e$ embedding matrix to obtain the vectorized representation for labels, resulting in the $m\times d_e$ label-to-vector matrix, where $d_e$ represents the embedding dimension (e.g., $d_e=128$ in our experiments). We also have the object style latent codes $Z_{obj}$ which is a $m\times d_{noise}$ noise matrix (e.g., $d_{noise}=128$ the same as $z_{img}$). We then concatenate the label-to-vector matrix and the noise matrix as the final $m\times (d_e+d_{noise})$ embedding matrix. So, the object instance style will depends on both the label embedding (semantics) and i.i.d. latent code (accounting for style variations). 

\textit{ii) Object instance-specific projection.} With the final embedding matrix, we compute object instance-specific channel-wise $\gamma$ and $\beta$ via linear projection with a learnable $(d_e+d_{noise})\times 2C$ projection matrix where $C$ is the number of channels.

\textit{iii) Mask prediction.} The $s\times s$ mask for each object instance (e.g., $s=16$ in our experiments) is predicted by a sub-network consisting of several up-sample convolution followed by sigmoid transformation. So, our predicted masks are not binary. Then, we resize the predicted masks to the sizes of corresponding bounding boxes.

\textit{iv) ISLA $\gamma$ and $\beta$ computation.} We unsqueeze the object instance-specific channel-wise $\gamma$ and $\beta$ to their corresponding bounding boxes with the predicted mask weights multiplied. Then, we add them together with averaged sum used for overlapping regions. 


\subsubsection{The Discriminator}
As shown in Figure~\ref{fig:workflow}, our discriminator consists of three components: the shared ResNet backbone, the image head classifier and the object head classifier. 

The ResNet backbone has several ResBlocks (4 for 64$\times$64 and 5 for 128$\times$128) as in \cite{miyato2018cgans,brock2018large}. The image head classifier consists of a ResBlock, a global average pooling layer and a fully-connected (FC) layer with one output unit, while object head classifier consists of ROI Align \cite{he2017mask}, a global average pooling layer and a FC layer with one output unit. 

Following the projection-based cGANs~\cite{miyato2018cgans} and the practice in BigGANs~\cite{brock2018large}, we learn a separate label embedding for computing object adversarial hinge loss.    

Denote by $D(\cdot;\Theta_D)$ the discriminator with parameters $\Theta_D$. Given an image $I$ (real or synthesized) and a layout $L$, the discriminator computes the prediction score for image and the average score for cropped objects, and we have, 
\begin{equation}
    (s_{img}, s_{obj}) = D(I, L;\Theta_D)
\end{equation}

\begin{figure*}[!ht]
    \centering
    \includegraphics[width=1.0\linewidth]{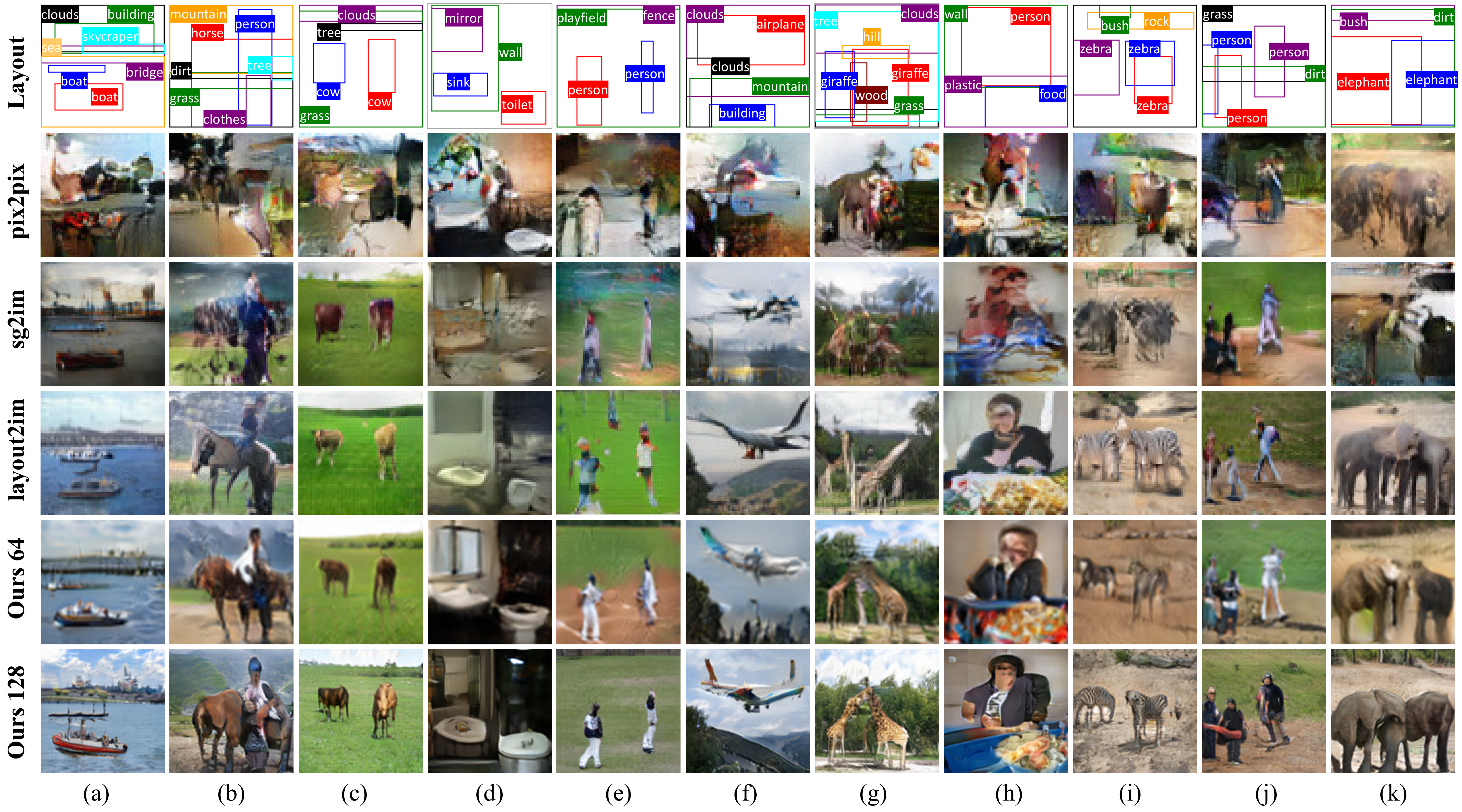}
    \includegraphics[width=1.0\linewidth]{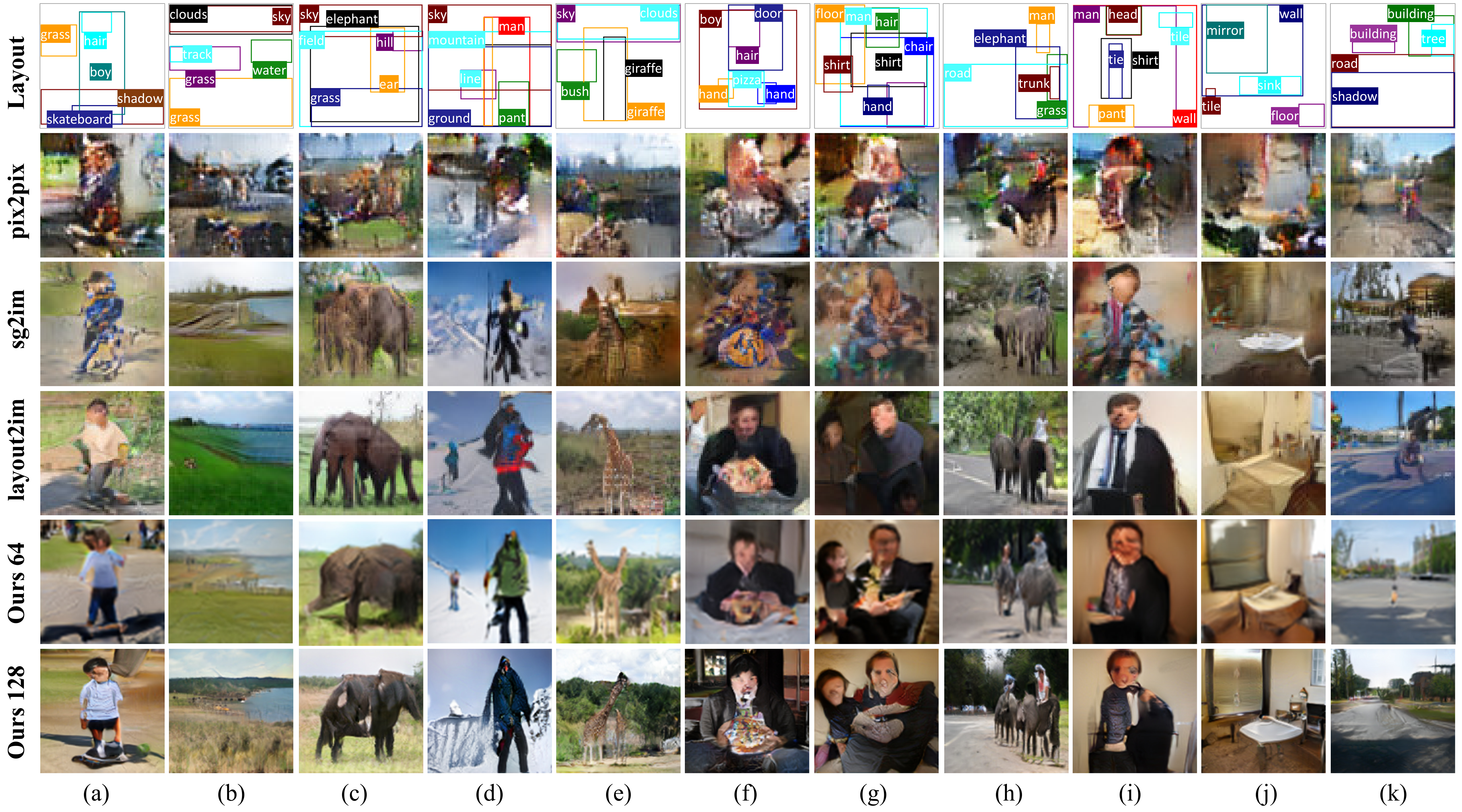}
    \caption{Generated samples from given layouts on COCO-Stuff (top) and Visual Genome (bottom). Images generated by pix2pix, sg2im, and layout2im are at 64$\times$64 resolution.}
    \label{fig:model_comparision}\vspace{-4mm}
\end{figure*}

\begin{table*}[!ht]
    \centering
    \resizebox{0.8\textwidth}{!}{
    \begin{tabular}{c|cc|cc|cc}
    \multirow{2}{*}{\textbf{Methods}} & \multicolumn{2}{c}{\textbf{Inception Score}} & \multicolumn{2}{|c}{\textbf{FID}} & \multicolumn{2}{|c}{\textbf{Diversity Score}} \\
    & \textbf{COCO} & \textbf{VG} & \textbf{COCO} & \textbf{VG} \\ \hline
    Real Images (64$\times$64) & 16.3 $\pm$ 0.4 & 13.9 $\pm$ 0.5 & - & - & - & - \\
    Real Images (128$\times$128) & 22.3 $\pm$ 0.5 & 20.5 $\pm$ 1.5 & - & - & - & - \\\hline
    pix2pix & 3.5 $\pm$ 0.1 & 2.7 $\pm$ 0.02 &121.97 &142.86 & 0 & 0 \\
    sg2im(GT Layout) & 7.3 $\pm$ 0.1 & 6.3 $\pm$ 0.2 &67.96 &74.61 & 0.02 $\pm$ 0.01 & 0.15 $\pm$ 0.12 \\
    Layout2Im & 9.1 $\pm$ 0.1 & 8.1 $\pm$ 0.1 & 38.14 & 40.07 & 0.15 $\pm$ 0.06 & 0.17 $\pm$ 0.09 \\ 
    Ours 64$\times$64 & \textbf{9.8 $\pm$ 0.2} & \textbf{8.7 $\pm$ 0.4} & \textbf{34.31} & \textbf{34.75} & \textbf{0.35 $\pm$ 0.09} & \textbf{0.34 $\pm$ 0.10} \\ \hline
    Ours 128$\times$128 & \textbf{13.8 $\pm$ 0.4} & \textbf{11.1 $\pm$ 0.6} & \textbf{29.65} & \textbf{29.36} & \textbf{0.40 $\pm$ 0.09} & \textbf{0.43 $\pm$ 0.09} \\ 
    \end{tabular}}
    \caption{Quantitative comparisons using Inception Score (higher is better), FID (lower is better) and Diversity Score (higher is better) evaluation on COCO-Stuff and VG dataset. Images for pix2pix~\cite{isola2017image}, sg2im~\cite{johnson2018image} and Layout2Im~\cite{zhao2018image} are at 64$\times$64 resolution.}
    \label{tab:eval_results} \vspace{-2mm}
\end{table*}

\subsubsection{The Loss Functions}\vspace{-1mm}
To train $(\Theta_G, \Theta_D)$ in our LostGAN, we utilize the hinge version~\cite{HingeLoss1,HingeLoss2} of the standard adversarial loss~\cite{goodfellow2014generative}, 
\begin{equation}
    l_{t}(I, L) = \begin{cases} \min (0, -1 + s_t);\, & \text{if } I \text{ is real} \\
    \min (0, -1 - s_t);\, & \text{if } I \text{ is fake} \end{cases}
\end{equation}
where $t\in \{img, obj\}$. Let $l(I,L)=l_{img}(I,L)+\lambda\cdot l_{obj}(I,L)$ with $\lambda$ the trade-off parameter for controlling the quality between synthesized images and objects ($\lambda=1$ in our experiments). We have the expected losses for the discriminator and the generator,
\begin{equation}
\begin{split}
\mathcal{L}(\Theta_D | \Theta_G) =& - \underset{(I,L) \sim p(I,L)}{\mathbb{E}}[l(I, L)] \\ 
\mathcal{L}(\Theta_G | \Theta_D) =& - \underset{(I,L)\sim p_{fake}(I,L)}{\mathbb{E}}[D(I, L; \Theta_D)]
\end{split}
\end{equation}
where $p(I, L)$ represents all the real and fake (by the current generator) data and $p_{fake}(I,L)$ represents the fake data. 

\section{Experiments}\vspace{-2mm}
 We test our LostGAN in the COCO-Stuff dataset \cite{caesar2018coco} and the Visual Genome (VG) dataset \cite{krishna2017visual}. We evaluate it for generating images at two resolutions 64$\times$64 and 128$\times$128. In comparison, the state-of-the-art methods include the very recent Layout2Im method~\cite{zhao2018image}, the scene graph to image (sg2im) method~\cite{johnson2018image} and the pix2pix method~\cite{isola2017image}.

\begin{figure*}[!ht]
    \centering
    \includegraphics[width=1\linewidth]{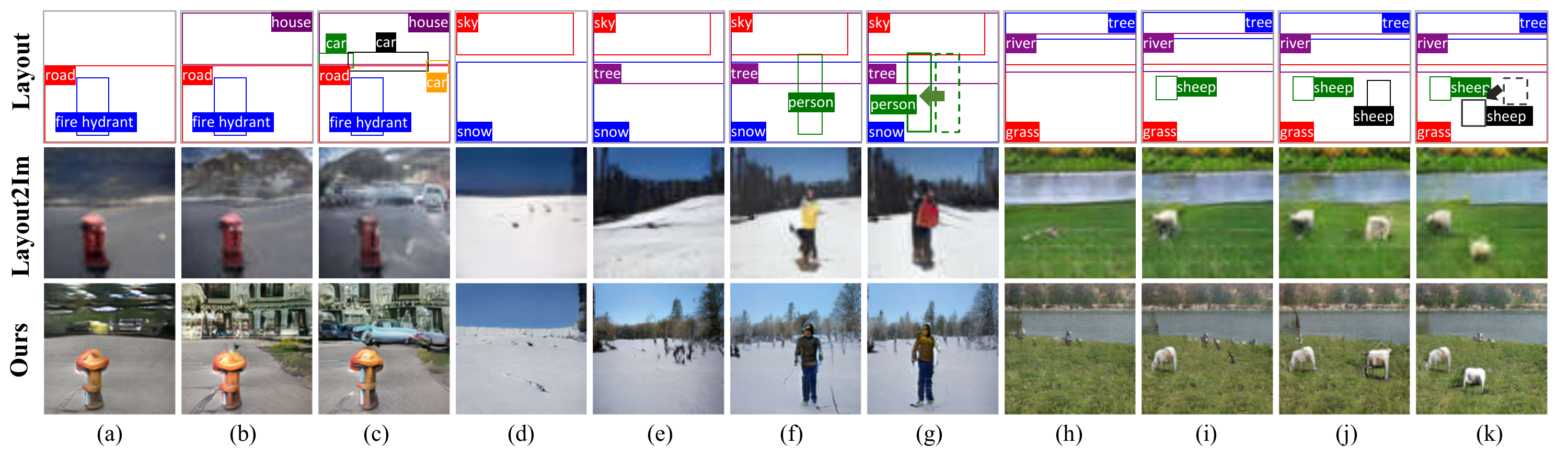}
    \caption{Generation results by adding new objects or change spatial position of objects.}
    \label{fig:reconfigrable_layout}\vspace{-2mm}
\end{figure*}

\begin{figure*}[!ht]
    \centering
    \includegraphics[width=1\linewidth]{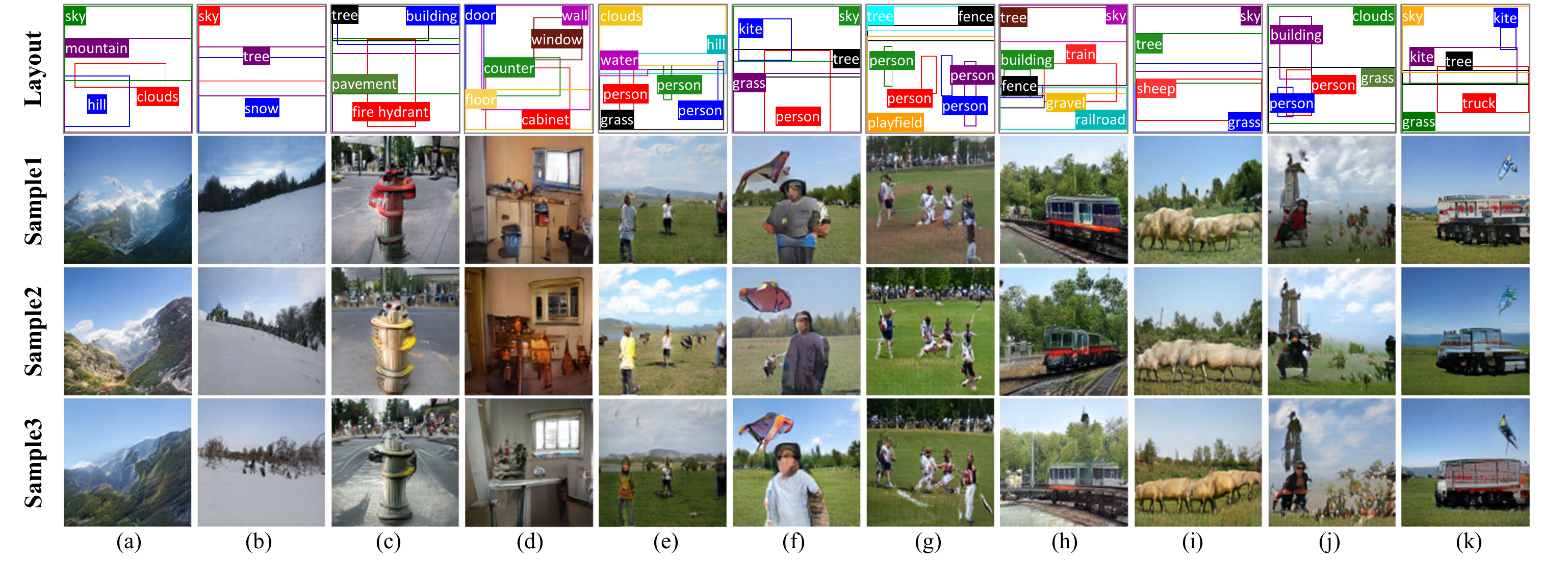}
    \caption{Multiple samples generated from same layout. Synthesized images have various visual appearance while preserving objects at desired location.}
    \label{fig:various_outs}\vspace{-2mm}
\end{figure*}

\begin{figure*}[!ht]
    \centering
    \includegraphics[width=1\linewidth]{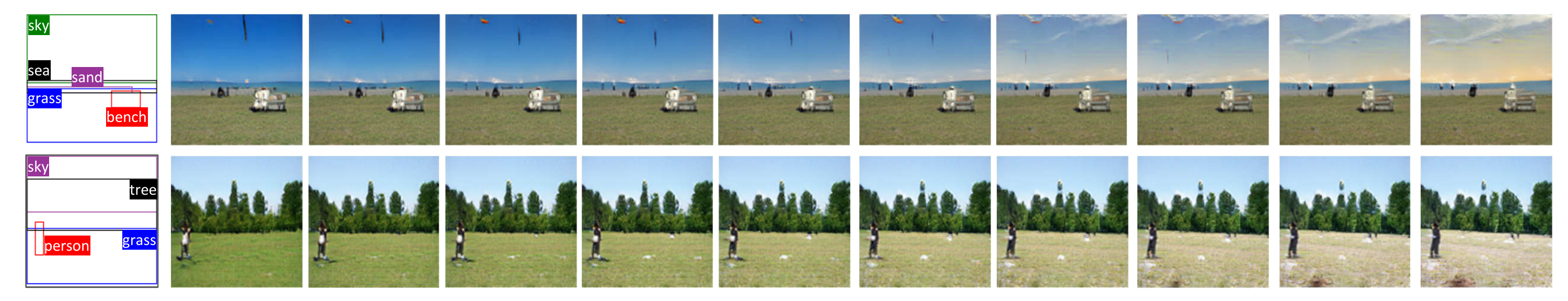}
    \caption{Linear interpolation of instance style. Top row indicates interpolation of style in sky, bottom row shows style morphing of grass.}
    \label{fig:instance_style}\vspace{-2mm}
\end{figure*}

\begin{figure*}[!ht]
    \centering
    \includegraphics[width=1\linewidth]{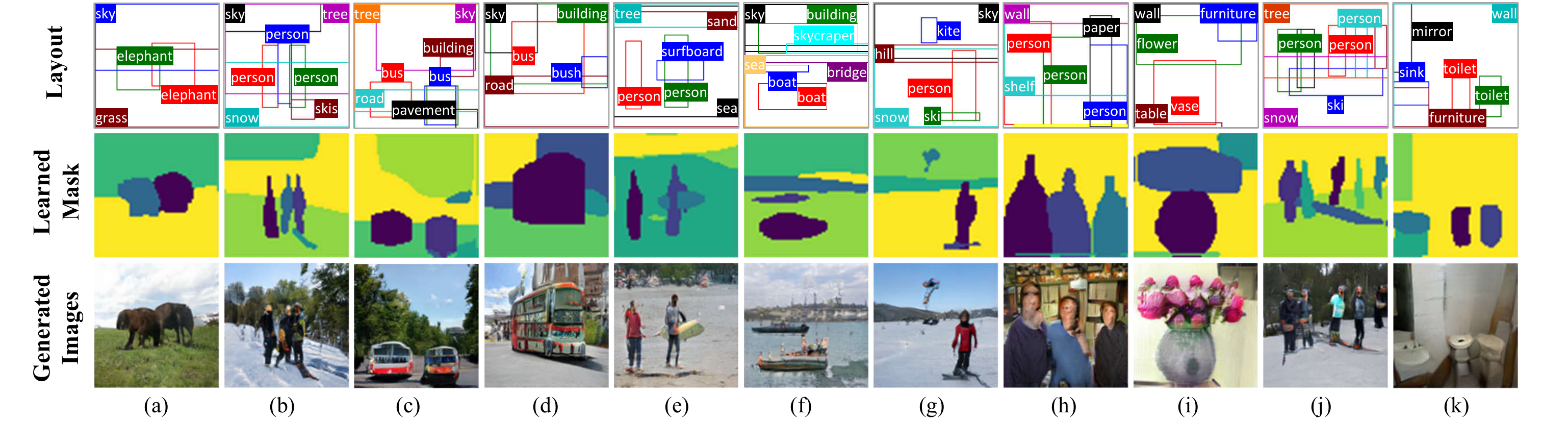}
    \caption{Synthesized images and learned masks for given layouts. Our proposed model learns masks from given layout in a weakly-supervised manner as ground truth mask for each object is not utilized during training.}
    \label{fig:generated_masks}\vspace{-2mm}
\end{figure*}

\subsection{Datasets}\vspace{-2mm}
The \textbf{COCO-Stuff} 2017 \cite{caesar2018coco} augments the COCO dataset with pixel-level stuff annotations. The annotation contains 80 \textit{thing} classes (person, car, \textit{etc.}) and 91 \textit{stuff} classes (sky, road, \textit{etc.})
Following settings of \cite{johnson2018image}, objects covering less than 2\% of the image are ignored, and we use images with 3 to 8 objects. 
The \textbf{Visual Genome} dataset \cite{krishna2017visual}. Following settings of \cite{johnson2018image} to  removing small and infrequent objects, we have 62,565 training, 5,506 val and 5,088 testing images with 3 to 30 objects from 178 categories in each image.

\vspace{-1mm}
\subsection{Evaluation Metrics}\vspace{-2mm}
We evaluate quality and visual appearance of generated images by Inception Score (higher is better) \cite{salimans2016improved} and Fr\`echet Inception Distance (FID, lower is better) \cite{heusel2017gans}, which use pretrained Inception \cite{szegedy2015going} network to encourage recognizable objects within images and diversity across images. Diversity score computes perceptual similarity between two images (higher is better). We adopt LPIPS metric \cite{zhang2018unreasonable} to compute perceptual similarity in feature space between two images generated from same layout as diversity score. We also evaluate our model by recently proposed Classification Accuracy Score (CAS) \cite{ravuri2019classification}.

\begin{table}[]
    \centering
    \begin{tabular}{c|cc}
    \hline
       \multirow{2}{*}{\textbf{Methods}} & \multicolumn{2}{c}{\textbf{Classification Accuracy}} \\
         & COCO & VG \\
        \hline
        Layout2im & 27.32 & 23.25\\
        Ours 64x64 & 28.81 & 27.50 \\
        Ours 128x128 & 28.70 & 25.89 \\
        Real Images & 51.04 & 48.07 \\

    \end{tabular}
    \caption{Classification Accuracy Comparisons. We train resnet-101 on cropped objects from generated images (generate five samples for each layout) and evaluate on objects from real images.}
    \label{tab:classification_score} \vspace{-4mm}
\end{table}{}

\vspace{-1mm}
\subsection{Quantitative results}\vspace{-2mm}
Table~\ref{tab:eval_results},~\ref{tab:classification_score} summarizes comparisons between our model and state-of-the-art models with respect to inception score, FID, diversity score and classification accuracy. Our LostGAN outperforms the most recent Layout2Im~\cite{zhao2018image} in terms of both Inception score and Diversity score. For 64$\times$64 images, the improvement of Inception score, FID and classification accuracy indicates higher visual quality of image generated by our model. Diversity score is improved significantly which shows that our LostGAN can generate images with various appearance for a given layout. We also conduct experiments at the resolution of 128$\times$128, and our LostGAN obtains consistently better results.

\vspace{-1mm}
\subsection{Qualitative results}\vspace{-2mm}
Figure~\ref{fig:model_comparision} shows results of different models generating images from the same layout on both COCO-Stuff and VG. The input layouts are quite complex. Our LostGAN can generate visually more appealing images with more recognizable objects that are consistent with input layouts at resolution 64$\times$64, and is further capable of synthesizing images at $128\times 128$ resolution with better image quality. 

We also conduct some ablation studies on the three aspects of reconfigurability and mask prediction.

\textbf{Layout reconfiguration} is demonstrated by adding object to or moving a bounding box in a layout (Figure~\ref{fig:reconfigrable_layout}). Our LostGAN shows better layout reconfigurability than the Layout2Im~\cite{zhao2018image}. When adding extra objects or moving bounding box of one instance, our model can generate reasonable objects at desired position while keeping existing objects unchanged as we keep the input style of existing objects fixed. When moving bounding box of one object, style of generated object in new position can also be kept consistent, like (f) and (g), the person is moved while keep style feature like pose and color of clothes unaffected. 

\textbf{Image style reconfiguration} To assess diversity of generation, multiple images are sampled from our LostGAN for each input layout (Figure~\ref{fig:various_outs}). Our model can synthesize images with different visual appearance for a given layout while preserving objects at desired location. 

\textbf{Object instance style reconfiguration} Our LostGAN is also capable of controlling styles at object instance level. Figure~\ref{fig:instance_style} shows results of gradually morphing styles of one instance in different images. Top row shows how the style of sky gradually turns from blue to dusk while keeping styles of other objects unaltered. Bottom row displays how the style of grass transforms from green to withered. 

\textbf{Weakly-supervised mask prediction} Figure~\ref{fig:generated_masks} shows generated semantic label map when synthesizing images from given layouts. For pixels where bounding boxes of different objects overlap, their semantic labels are assigned by objects with the highest predicted mask weight. Unlike \cite{hong2018inferring,li2019object} where ground truth masks is adopted to guide learning of shape generator, our model can learn semantic masks in a weakly-supervised manner. Even for objects with overlapped bounding box, like person and surfboard in (f), synthesized images and learned masks are consistent and semantically reasonable.

\vspace{-1mm}
\section{Conclusion}\vspace{-2mm}
This paper presents a layout- and style-based architecture for generative adversarial networks (LostGANs) that can be trained end-to-end to generate images from reconfigurable layout and style. The proposed LostGAN can learn fine-grained mask maps in a weakly-supervised manner to bridge the gap between layouts and images, and proposes the object instance-specific layout-aware feature normalization (ISLA-Norm) in the generator to realize multi-object style generation. State-of-the-art performance is obtained on COCO-Stuff and VG dataset. Qualitative results demonstrate the proposed model is capable of generating scene images with reconfigurable layout and instance-level style control. 

\vspace{-1mm}
\section*{Acknowledgement}
\vspace{-1mm}
The authors would like to thank the anonymous reviewers for their helpful comments. This work was supported in part by ARO grant W911NF1810295, NSF IIS-1909644, Salesforce Inaugural Deep Learning Research Grant (2018) and ARO DURIP grant W911NF1810209. The views presented in this paper are those of the authors and should not be interpreted as representing any funding agencies.

{\small
\bibliographystyle{ieee_fullname}
\bibliography{egbib}

\begin{thebibliography}{10}\itemsep=-1pt

\bibitem{brock2018large}
Andrew Brock, Jeff Donahue, and Karen Simonyan.
\newblock Large scale gan training for high fidelity natural image synthesis.
\newblock {\em arXiv preprint arXiv:1809.11096}, 2018.

\bibitem{caesar2018coco}
Holger Caesar, Jasper Uijlings, and Vittorio Ferrari.
\newblock Coco-stuff: Thing and stuff classes in context.
\newblock In {\em Proceedings of the IEEE Conference on Computer Vision and
  Pattern Recognition}, pages 1209--1218, 2018.

\bibitem{de2017modulating}
Harm De~Vries, Florian Strub, J{\'e}r{\'e}mie Mary, Hugo Larochelle, Olivier
  Pietquin, and Aaron~C Courville.
\newblock Modulating early visual processing by language.
\newblock In {\em Advances in Neural Information Processing Systems}, pages
  6594--6604, 2017.

\bibitem{dumoulin2016adversarially}
Vincent Dumoulin, Ishmael Belghazi, Ben Poole, Olivier Mastropietro, Alex Lamb,
  Martin Arjovsky, and Aaron Courville.
\newblock Adversarially learned inference.
\newblock {\em arXiv preprint arXiv:1606.00704}, 2016.

\bibitem{dumoulin2017learned}
Vincent Dumoulin, Jonathon Shlens, and Manjunath Kudlur.
\newblock A learned representation for artistic style.
\newblock {\em Proc. of ICLR}, 2, 2017.

\bibitem{goodfellow2014generative}
Ian Goodfellow, Jean Pouget-Abadie, Mehdi Mirza, Bing Xu, David Warde-Farley,
  Sherjil Ozair, Aaron Courville, and Yoshua Bengio.
\newblock Generative adversarial nets.
\newblock In {\em Advances in neural information processing systems}, pages
  2672--2680, 2014.

\bibitem{he2017mask}
Kaiming He, Georgia Gkioxari, Piotr Doll{\'a}r, and Ross Girshick.
\newblock Mask r-cnn.
\newblock In {\em Proceedings of the IEEE international conference on computer
  vision}, pages 2961--2969, 2017.

\bibitem{he2016deep}
Kaiming He, Xiangyu Zhang, Shaoqing Ren, and Jian Sun.
\newblock Deep residual learning for image recognition.
\newblock In {\em Proceedings of the IEEE conference on computer vision and
  pattern recognition}, pages 770--778, 2016.

\bibitem{FID}
Martin Heusel, Hubert Ramsauer, Thomas Unterthiner, Bernhard Nessler, and Sepp
  Hochreiter.
\newblock {GANs} trained by a two time-scale update rule converge to a local
  nash equilibrium.
\newblock In {\em NIPS}, 2017.

\bibitem{heusel2017gans}
Martin Heusel, Hubert Ramsauer, Thomas Unterthiner, Bernhard Nessler, and Sepp
  Hochreiter.
\newblock Gans trained by a two time-scale update rule converge to a local nash
  equilibrium.
\newblock In {\em Advances in Neural Information Processing Systems}, pages
  6626--6637, 2017.

\bibitem{hinz2019generating}
Tobias Hinz, Stefan Heinrich, and Stefan Wermter.
\newblock Generating multiple objects at spatially distinct locations.
\newblock {\em arXiv preprint arXiv:1901.00686}, 2019.

\bibitem{hong2018inferring}
Seunghoon Hong, Dingdong Yang, Jongwook Choi, and Honglak Lee.
\newblock Inferring semantic layout for hierarchical text-to-image synthesis.
\newblock In {\em Proceedings of the IEEE Conference on Computer Vision and
  Pattern Recognition}, pages 7986--7994, 2018.

\bibitem{huang2018multimodal}
Xun Huang, Ming-Yu Liu, Serge Belongie, and Jan Kautz.
\newblock Multimodal unsupervised image-to-image translation.
\newblock In {\em Proceedings of the European Conference on Computer Vision
  (ECCV)}, pages 172--189, 2018.

\bibitem{ioffe2015batch}
Sergey Ioffe and Christian Szegedy.
\newblock Batch normalization: Accelerating deep network training by reducing
  internal covariate shift.
\newblock {\em arXiv preprint arXiv:1502.03167}, 2015.

\bibitem{isola2017image}
Phillip Isola, Jun-Yan Zhu, Tinghui Zhou, and Alexei~A Efros.
\newblock Image-to-image translation with conditional adversarial networks.
\newblock In {\em Proceedings of the IEEE conference on computer vision and
  pattern recognition}, pages 1125--1134, 2017.

\bibitem{johnson2018image}
Justin Johnson, Agrim Gupta, and Li Fei-Fei.
\newblock Image generation from scene graphs.
\newblock In {\em Proceedings of the IEEE Conference on Computer Vision and
  Pattern Recognition}, pages 1219--1228, 2018.

\bibitem{karras2017progressive}
Tero Karras, Timo Aila, Samuli Laine, and Jaakko Lehtinen.
\newblock Progressive growing of gans for improved quality, stability, and
  variation.
\newblock {\em arXiv preprint arXiv:1710.10196}, 2017.

\bibitem{karras2018style}
Tero Karras, Samuli Laine, and Timo Aila.
\newblock A style-based generator architecture for generative adversarial
  networks.
\newblock {\em arXiv preprint arXiv:1812.04948}, 2018.

\bibitem{kim2017learning}
Taeksoo Kim, Moonsu Cha, Hyunsoo Kim, Jung~Kwon Lee, and Jiwon Kim.
\newblock Learning to discover cross-domain relations with generative
  adversarial networks.
\newblock In {\em Proceedings of the 34th International Conference on Machine
  Learning-Volume 70}, pages 1857--1865. JMLR. org, 2017.

\bibitem{krishna2017visual}
Ranjay Krishna, Yuke Zhu, Oliver Groth, Justin Johnson, Kenji Hata, Joshua
  Kravitz, Stephanie Chen, Yannis Kalantidis, Li-Jia Li, David~A Shamma, et~al.
\newblock Visual genome: Connecting language and vision using crowdsourced
  dense image annotations.
\newblock {\em International Journal of Computer Vision}, 123(1):32--73, 2017.

\bibitem{li2019object}
Wenbo Li, Pengchuan Zhang, Lei Zhang, Qiuyuan Huang, Xiaodong He, Siwei Lyu,
  and Jianfeng Gao.
\newblock Object-driven text-to-image synthesis via adversarial training.
\newblock {\em arXiv preprint arXiv:1902.10740}, 2019.

\bibitem{HingeLoss2}
Jae~Hyun Lim and Jong~Chul Ye.
\newblock Geometric gan.
\newblock {\em arXiv preprint arXiv:1705.02894}, 2017.

\bibitem{miyato2018spectral}
Takeru Miyato, Toshiki Kataoka, Masanori Koyama, and Yuichi Yoshida.
\newblock Spectral normalization for generative adversarial networks.
\newblock {\em arXiv preprint arXiv:1802.05957}, 2018.

\bibitem{miyato2018cgans}
Takeru Miyato and Masanori Koyama.
\newblock cgans with projection discriminator.
\newblock {\em arXiv preprint arXiv:1802.05637}, 2018.

\bibitem{odena2017conditional}
Augustus Odena, Christopher Olah, and Jonathon Shlens.
\newblock Conditional image synthesis with auxiliary classifier gans.
\newblock In {\em Proceedings of the 34th International Conference on Machine
  Learning-Volume 70}, pages 2642--2651. JMLR. org, 2017.

\bibitem{park2019semantic}
Taesung Park, Ming-Yu Liu, Ting-Chun Wang, and Jun-Yan Zhu.
\newblock Semantic image synthesis with spatially-adaptive normalization.
\newblock In {\em Proceedings of the IEEE Conference on Computer Vision and
  Pattern Recognition}, pages 2337--2346, 2019.

\bibitem{radford2015unsupervised}
Alec Radford, Luke Metz, and Soumith Chintala.
\newblock Unsupervised representation learning with deep convolutional
  generative adversarial networks.
\newblock {\em arXiv preprint arXiv:1511.06434}, 2015.

\bibitem{ravuri2019classification}
Suman Ravuri and Oriol Vinyals.
\newblock Classification accuracy score for conditional generative models.
\newblock {\em arXiv preprint arXiv:1905.10887}, 2019.

\bibitem{reed2016generative}
Scott Reed, Zeynep Akata, Xinchen Yan, Lajanugen Logeswaran, Bernt Schiele, and
  Honglak Lee.
\newblock Generative adversarial text to image synthesis.
\newblock {\em arXiv preprint arXiv:1605.05396}, 2016.

\bibitem{salimans2016improved}
Tim Salimans, Ian Goodfellow, Wojciech Zaremba, Vicki Cheung, Alec Radford, and
  Xi Chen.
\newblock Improved techniques for training gans.
\newblock In {\em Advances in neural information processing systems}, pages
  2234--2242, 2016.

\bibitem{szegedy2015going}
Christian Szegedy, Wei Liu, Yangqing Jia, Pierre Sermanet, Scott Reed, Dragomir
  Anguelov, Dumitru Erhan, Vincent Vanhoucke, and Andrew Rabinovich.
\newblock Going deeper with convolutions.
\newblock In {\em Proceedings of the IEEE conference on computer vision and
  pattern recognition}, pages 1--9, 2015.

\bibitem{HingeLoss1}
Dustin Tran, Rajesh Ranganath, and David~M Blei.
\newblock Deep and hierarchical implicit models.
\newblock {\em arXiv preprint arXiv:1702.08896}, 7, 2017.

\bibitem{wang2018high}
Ting-Chun Wang, Ming-Yu Liu, Jun-Yan Zhu, Andrew Tao, Jan Kautz, and Bryan
  Catanzaro.
\newblock High-resolution image synthesis and semantic manipulation with
  conditional gans.
\newblock In {\em Proceedings of the IEEE Conference on Computer Vision and
  Pattern Recognition}, pages 8798--8807, 2018.

\bibitem{xu2018attngan}
Tao Xu, Pengchuan Zhang, Qiuyuan Huang, Han Zhang, Zhe Gan, Xiaolei Huang, and
  Xiaodong He.
\newblock Attngan: Fine-grained text to image generation with attentional
  generative adversarial networks.
\newblock In {\em Proceedings of the IEEE Conference on Computer Vision and
  Pattern Recognition}, pages 1316--1324, 2018.

\bibitem{zhang2018self}
Han Zhang, Ian Goodfellow, Dimitris Metaxas, and Augustus Odena.
\newblock Self-attention generative adversarial networks.
\newblock {\em arXiv preprint arXiv:1805.08318}, 2018.

\bibitem{zhang2017stackgan}
Han Zhang, Tao Xu, Hongsheng Li, Shaoting Zhang, Xiaogang Wang, Xiaolei Huang,
  and Dimitris~N Metaxas.
\newblock Stackgan: Text to photo-realistic image synthesis with stacked
  generative adversarial networks.
\newblock In {\em Proceedings of the IEEE International Conference on Computer
  Vision}, pages 5907--5915, 2017.

\bibitem{zhang2018unreasonable}
Richard Zhang, Phillip Isola, Alexei~A Efros, Eli Shechtman, and Oliver Wang.
\newblock The unreasonable effectiveness of deep features as a perceptual
  metric.
\newblock In {\em Proceedings of the IEEE Conference on Computer Vision and
  Pattern Recognition}, pages 586--595, 2018.

\bibitem{zhao2018image}
Bo Zhao, Lili Meng, Weidong Yin, and Leonid Sigal.
\newblock Image generation from layout.
\newblock {\em arXiv preprint arXiv:1811.11389}, 2018.

\bibitem{zhu2017unpaired}
Jun-Yan Zhu, Taesung Park, Phillip Isola, and Alexei~A Efros.
\newblock Unpaired image-to-image translation using cycle-consistent
  adversarial networks.
\newblock In {\em Proceedings of the IEEE International Conference on Computer
  Vision}, pages 2223--2232, 2017.

\end{thebibliography}
}

\end{document}